\documentclass[letterpaper, 10 pt, conference]{ieeeconf}  % Comment this line out
\IEEEoverridecommandlockouts                              % This command is only
\usepackage{times}
\usepackage{latexsym}

\usepackage{CJKutf8}
\usepackage{amsmath}
\usepackage{amssymb}
\usepackage{multirow}
\usepackage{subcaption}
\usepackage{graphicx}

\usepackage{url}

\title{\LARGE \bf
CAN-NER: Convolutional Attention Network for \\Chinese Named Entity Recognition
}

\author{Yuying Zhu$^{1}$, Guoxin Wang$^{2}$, Börje F. Karlsson$^{3}$% <-this % stops a space \\
\thanks{$^{1}$ Yuying Zhu is from Nankai University, Tianjin, China
        {\tt\small yuyzhu@mail.nankai.edu.cn}}%
\thanks{$^{2}$ Guoxin Wang is from Microsoft Research Asia, Beijing, China
        {\tt\small guow@microsoft.com}}%
\thanks{$^{3}$ Börje F. Karlsson is from Microsoft Research Asia, Beijing, China
{\tt\small borjekar@microsoft.com}}%
}

\begin{document}

\maketitle

\begin{CJK}{UTF8}{gbsn}

\begin{abstract}
Named entity recognition (NER) in Chinese is essential but difficult because of the lack of natural delimiters.
Therefore, Chinese Word Segmentation (CWS) is usually considered as the first step for Chinese NER.
However, models based on word-level embeddings and lexicon features often suffer from segmentation errors and out-of-vocabulary (OOV) words.
In this paper, we investigate a \textbf{C}onvolutionall \textbf{A}ttention \textbf{N}etwork called \textbf{CAN} for Chinese NER, which consists of a character-based convolutional neural network (CNN) with local-attention layer and a gated recurrent unit (GRU) with global self-attention layer to capture the information from adjacent characters and sentence contexts.
Also, compared to other models, not depending on any external resources like lexicons and employing small size of char embeddings make our model more practical.
Extensive experimental results show that our approach outperforms state-of-the-art methods without word embedding and external lexicon resources
on different domain datasets including Weibo, MSRA and Chinese Resume NER dataset.
\end{abstract}

\section{Introduction}\label{Intro}
Named Entity Recognition (NER) aims at identifying text spans which are associated with a specific semantic entity type such as person (PER), organization (ORG), location (LOC), and geopolitical entity (GPE).
NER receives constant research attention as it is the first step in a wide range of downstream Natural Language Processing (NLP) tasks, e.g., entity linking~\cite{gupta2017entity}, relation extraction~\cite{miwa2016end}, event extraction~\cite{chen2015event}, and co-reference resolution~\cite{fragkou2017applying}.
%QA information extraction?
%The standard method of existing state-of-the-art models for English NER use word-based LSTM-CRF models \cite{lample2016neural,ma2016end,chiu2016named,AAAI1817123}.
%Most existing work for English NER treats it as a word-by-word sequence label task and makes full use of the RNN with the CRF to capture adjacent context information at the word-level.

The standard method of existing state-of-the-art models for English NER treats it as a word-by-word sequence labeling task and makes full use of the Recurrent Neural Network (RNN) and Conditional Random Field (CRF) to capture context information at the word level \cite{lample2016neural,ma2016end,chiu2016named,AAAI1817123}. 
%For example, \cite{lample2016neural} employ a long Short-Term Memory (LSTM) network with CRF to reason jointly over tagging decisions for each token and capture both orthographic evidence and distributional evidence.
%\cite{jiao2018chinese} construct a deep Bidirectional Gated Recurrent Unit (Bi-GRU) network with CRF for a joint lexical analysis task of word segmentation, part-of-speech (POS) tagging, and NER.
%重新看一下文献
\begin{figure}[htbp]
\small
\centering
\fbox{\parbox{0.6\linewidth}{
	\textbf{Sentence:} \\
	南京市长江大桥 \\
	\\
	\textbf{Segmentation 1:} \\
	南京市 $|$ 长江大桥 \\
	Nanjing City, Yangtze River Bridge \\
	Location, Location \\
	\\
	\textbf{Segmentation 2:} \\
	南京 $|$ 市长 $|$ 江大桥 \\
	Nanjing, Mayor, Jiang Daqiao \\
	Location, Title, Person
}}
	\caption{Entity Ambiguity with Word Segmentation.}
	\label{fg.wordseg}
\end{figure}
%Some Chinese NER work utilizes a tokenizer to get word boundaries and feeds them to the word-level sequence label model similar to English NER models (reference here, 没找到直接把分词作为entity boudary的).
%So, most Chinese NER works utilize a tokenizer to get word boundaries and feeds them to the word-level sequence label model similar to English NER models.
These models for English NER predict a tag for each word assuming that words can be separated clearly by explicit word separators, e.g., blank space.
Therefore, for Chinese language without natural delimiters, it is intuitive to apply Chinese Word Segmentation (CWS) first to get word boundaries and then use a word-level sequence labeling model similar to the English NER models.  
However, in Chinese language, word boundaries can be ambiguous, which leads the possibility that entity boundaries does not match word boundaries.
For example, in a sentence, ``西藏自治区 (Tibet Autonomous Region)" is a GPE type in NER task while can be segmented as one single word or as two words ``西藏 (Tibet)" and ``自治区 (autonomous region)" separately, depending on different granularity of segmentation tools.
But most of the time, it is hard to determine or unify the granularity of word segmentation.
Also, as shown in Figure~\ref{fg.wordseg}, different segmentation sometimes leads different sentence meaning in Chinese, which can even result in different named entities. 
%\bk{For Chinese NER task, most of recent models rely on the word-level embeddings and external lexicon sets. }
%It is a straightforward intuition to utilize a tokenizer to get word boundaries and feeds them to the word-level sequence label model similar to English NER models.
%However, it is hard to disambiguate with few context information in Chinese, that causes the accuracy of word segmentation to be lower.
%For example, as shown in Figure~\ref{fg.wordseg}, for this sentence, different segmentation could lead to dissimilar entity results.
%The segmentation 1 gets two locations: \textit{Nanjing City} and \textit{Yangtze River Bridge}, and \textit{Yangtze River Bride} is located in \textit{Nanjing City}.
%However, segmentation 2 gets a location, a title and a person, which means \textit{Daqiao Jiang} is the \textit{mayor} of \textit{Nanjing}.
%However, it is could also be meaningful to capture the words ``市长'' (mayor), a location \textit{Nanjing}, and a person named \textit{Jiang Daqiao}, respectively, which means \textit{Daqiao Jiang} is the \textit{mayor} of \textit{Nanjing City}.
Obviously, it is impossible to make the right extraction with word-based NER model if the boundaries are mistakenly detected at the first time.
%Obviously, it is impossible to make the right extraction in the word-based NER model if the boundaries between a term and its contexts are mistakenly detected.
Most recent neural network based Chinese NER models rely heavily on the word-level embeddings and external lexicon sets \cite{huang2017addressing,zhang2018chinese}.
%Recent neural network based Chinese NER models are heavily dependent on the word embedding and lexicon feature \cite{huang2017addressing,zhang2018chinese}.
%The quality of those models would potentially be affected by different word embedding representations and lexicon features, and it is also susceptible to the out-of-vocabulary (OOV) issue.
The quality of those models would potentially be affected by different word embedding representations and lexicon features.
Moreover, word-based models will suffer from OOV issues for Chinese words can be enormous and named entities are important source of OOV words.
We also list other potential problems as follows:
(1) Word embeddings dependency increases the model size and makes the fine-tuning process harder in the training step;
(2) It is hard to learn word representation correctly without enough labeled utterances for named entities are usually the proper nouns.
(3) Large lexicons are much expensive for real NER system because it will spend large memories and long matching time to obtain the features, that makes the model inefficient.
(4) It is barely impossible to clean noise words in the large lexicon. Both of word embeddings and lexicon are hard to be updated after trained.

%\bk{These large lookup tables make model inefficient and inconstancy both in the process of fine-tuning and predicting as while.
%\item1. Word embeddings dependency increases the model size and makes the fine-tuning process harder in the training step. 
%\item2. Name entities usually are the rare words, so it is hard to learn word representation correctly without enough labeled utterances.
%\item3. Large lexicons are much expensive for real NER system because it will spend large memories and long matching time to obtain the features, that makes the model inefficient.
%\item4. It is barely impossible to clean noise words in the large lexicon. Both of word embeddings and lexicon are hard to be updated after trained.}

% \begin{itemize}
% \item It is costly to construct word embeddings and lexicons of high quality, and is almost impossible to filter noise words in the large lexicons. 
% %It is hardly to re-produce with private corpus data for some recent embedding models, %like BERT, .
% \item It is hardly to learn word representations correctly without enough labeled utterances for named entities are usually the proper nouns.
% \item Large lexicons are much expensive for real NER system because it will spend more memories and matching time to generate features which makes the model inefficient. 
% \end{itemize}
%? do we need? 5. It is difficult to re-produce the embedding model result with private corpus, like BERT \cite{devlin2018bert}.

% To avoid these obstacles, we propose a character-based NER model.
Moreover, character embedding can only carry limited information for losing word and word sequence information.
For instance, the character ``拍'' in word ``球拍'' (bat) and ``拍卖'' (auction) has entirely different meanings.
How to better integrate the segmentation information and exploit local context information is the key in the character-based model.
%Moreover, character embedding can only carry limited information compared with explicit word-level and sentence-level structures.
%For instance, the character ``拍'' in word ``球拍'' (bat) and ``拍卖'' (auction) have entirely different meanings.
\cite{zhang2018chinese} leverage lexicons to add all the embeddings of candidate word segmentation to their last character embeddings as soft features and construct a convolutional neural network (CNN) to encode characters as word-level information.
\cite{cao2018adversarial} propose a multi-task architecture to learn NER tagging and Chinese word segmentation together with each part using a character-based Bi-LSTM.
%Therefore, character-based neural networks are employed to acquire character-level relations instead of directly detecting entity information by word segmentation.
We propose a convolutional attention layer to extract the implicit local context features from character sequence. 
With the segmentation vector softly concatenating into character embedding, the convolutional attention layer is able to group implicitly meaning-related characters and reduce the impact of segmentation errors. 
Results show that our model outperforms other Chinese NER models without external resources.

%In this paper, we propose a local attention layer inside the window of contexts before feeding them into the CNN layer to capture the implicit relations within adjacent characters, in which the position features from word segmentation are soft hints for character combinations.
%After the CNN layer, we conduct a BiGRU structure with a global self-attention layer on the whole sentence to gain the sentence-level dependencies for self-attention is proved to be effective in Chinese NER task \cite{cao2018adversarial,tan2018deep}.
%The final CRF layer is used as the decoder to label each character.
%In experiments on %\bk{three}
%four datasets, we show that our model outperforms purely character-based and word-based neural models which rely on external word embedding and lexicon resources.
The main contributions of this paper can be summarized as follows:
\begin{itemize}
    \item We first combine the CNN with the local-attention mechanism to enhance the ability of the model to capture implicitly local context relations among character sequence. Compared with experimental results of baseline with normal CNN layer, our Convolutional Attention layer leads a remarkable improvement of F1 performance.
    %\item We first combine the CNN with the local-attention mechanism to enhance the ability of capturing relations between Chinese characters and their local contexts. Compared with experimental results of baseline with normal CNN layer, our Convolutional Attention layer leads a significant improvement of F1 performance.
	%\item We modify the CNN layer with the local-attention mechanism to enhance the ability of capturing Chinese character meanings in local contexts. Comparing with experiment result of baseline with CNN layer, our Convolutional Attention layer leads a significant improvement of F1-score.
	\item We introduce a character-based Chinese NER model that consists of CNN with local attention and Bi-GRU with global self-attention layers. Our model achieve state-of-the-art F1-scores without using any external word embeddings and lexicon resources, which is more practical for real NER system. 
	%\item We introduce a character-based Chinese NER model that consists of CNN with local attention and GRU with global self-attention layers. Our model outperforms previous work on Weibo and Chinese Resume datasets without any external word embeddings and lexicon resources dependencies. 
\end{itemize}

\begin{figure}[tb]
	\centering
	\includegraphics[width=\linewidth]{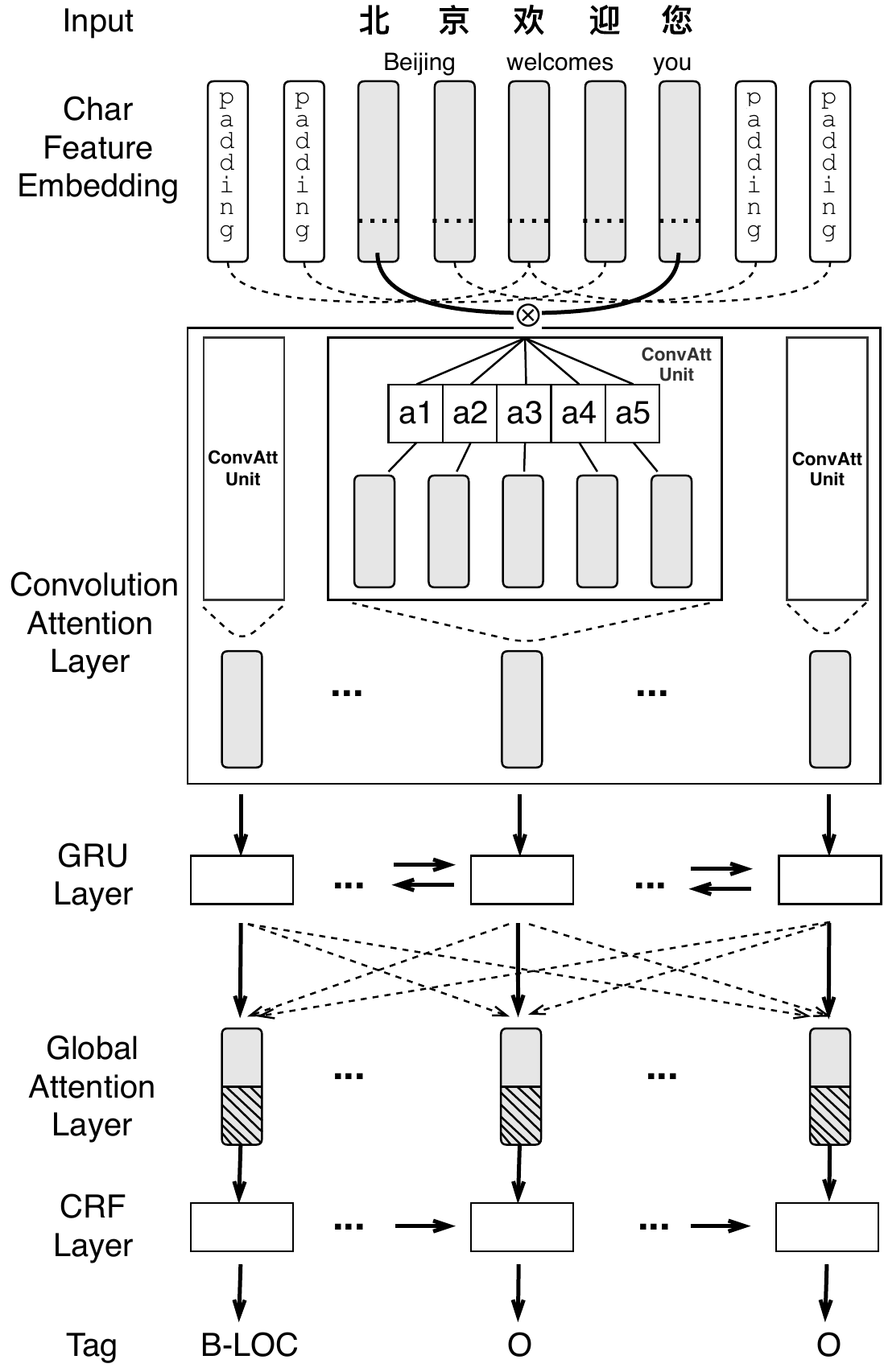}
	\caption{The Whole Architecture.
			A convolutional attention layer is constructed to encode both character- and word-level information.
			The BiGRU-CRF layer is extended by a global self-attention layer to capture long sequential sentence-level relations.
			}
	\label{fg.architecture}
\end{figure}

\section{Methodology}
We utilize BiGRU-CRF as our basic structure.
% following state-of-the-art English models (Huang et al., 2015; Ma and Hovy, 2016; Lample et al., 2016). (改成ref格式，之前写的LSTM-CRF，这里要换几个reference)
Our model considers multi-level context features using convolutional attention layer, GRU layer and global attention layer.
%CNN layer and attention mechanism are employed to generate multi-level information features.
The whole architecture of our proposed model is illustrated in Figure~\ref{fg.architecture}.

\subsection{Formulation}
%In the Chinese NER task, We denote an input sentence as $ \mathbf{X}_{i} = \{x_{i,1}, x_{i,2}, x_{i,3}, ..., x_{i,|\mathbf{X}_i|}\} $, where $ x_{i,j} \in \mathbb{R}^{d_e} $ represents the $j$-th character in sentence $\mathbf{X}_i$ and $d_e$ is the input embedding dimension.
%The whole corpus $ \mathcal{D} = \{(\mathbf{X}_i, \mathbf{Y}_i)\}_{i=1}^K $ consists of all $K$ sentences and label pairs $ (\mathbf{X}_i, \mathbf{Y}_i) $ where $ \mathbf{Y}_i = \{y_{i,1}, y_{i,2}, y_{i,3}, ..., y_{i,|\mathbf{Y}_i|}\} $ and $ y_{i,j} \in \mathcal{Y} $ belonging to the set of all entity types with a special label `\textit{O}' which indicates the character is not a part of entity.
In the Chinese NER task, we denote an input sentence as $ \mathbf{X}_{i} = \{x_{i,1}, x_{i,2}, x_{i,3}, ..., x_{i,\tau}\} $, where $ x_{i,\tau} \in \mathbb{R}^{d_e} $ represents the $\tau$-th character in sentence $\mathbf{X}_i$ and $d_e$ is the dimension of the input embeddings.
Correspondingly, we denote the sentence label sequence as $ \mathbf{Y}_i = \{y_{i,1}, y_{i,2}, y_{i,3}, ..., y_{i,\tau}\} $, where $ y_{i,\tau} \in \mathcal{Y} $ belongs to the set of all possible labels.
The objective is learning a function $ f_{\theta}: \mathbf{X} \mapsto \mathbf{Y} $ to obtain the entity types including the `\textit{O}' type for all the characters in the input text.
In the following text, we take one instance as the example and therefore omitting subindex $i$ in the formula.
%We define the model taking one instance as an example and omit subindex $i$ in later.

%\subsection{CNN Featurizer with Local Attention}
\subsection{Convolutional Attention Layer}
The convolutional attention layer aims to encode the sequence of input characters and implicitly group meaning-related characters in the local context.

The input representation for each character is constructed as $ x = [x_{ch}; x_{seg}] $, where $ x_{ch} \in \mathbb{R}^{d_{ch}} $ and $ x_{seg} \in \mathbb{R}^{d_{seg}} $ are character embedding and segmentation mask, respectively.
The segmentation information is encoded by BMES scheme \cite{wang2017convolutional}.
%in which the segmentation mask is represented by BMES scheme.
%The input representation is constructed as $ x = [x_{ch}; x_{pos}; x_{seg}; x_{cl}] $ where $ x_{ch} \in \mathbb{R}^{d_{ch}} $, $ x_{pos} \in \mathbb{R}^{d_{pos}} $, $ x_{seg} \in \mathbb{R}^{d_{seg}} $, and $ x_{cl} \in \mathbb{R}^{d_{cl}} $ are character embedding, position information, segmentation mask, and cluster features,
%(position information应该放到local-attention cnn处,cluster features我们如果没有用到应该删除？closelist feature需要提吗？)
%respectively, and the input embedding dimension $ d_e = d_{ch} + d_{pos} + d_{seg} + d_{cl} $.
%The character embedding provides character representation vector and 
%segmentation mask presents text segmentation information by BMES scheme .
%The position embedding distinguish the character order within a context window and the cluster features aim at grouping the similar characters in order to reduce the diversity of Chinese characters.

%We propose a convolutional attention layer to encode both character-level %and word-level features from input embeddings.
For every window in CNN, whose window size is $k$, we first concatenate a position embedding to each character embedding, helping to keep the sequential relations in the local window context.
The dimension of the position embedding equals to the window size $k$ with the initial values of $1$ at the position where the character lies in the window and $0$ at other positions.
So, the dimension of the concatenated embedding is $ d_e = d_{ch} + d_{pos} + d_{seg} $.
Then we apply a local attention inside the window to capture the relations between the center character and each context token, followed by a CNN with sum pooling layer.
%For a window size $k$, we first concatenate a position embedding to each character to keep the sequential relation between them and the input embedding dimension $ d_e = d_{ch} + d_{pos} + d_{seg} $.
%Then we apply a local attention among the window to capture the relations between the center character and each context token, followed by a CNN with sum pooling layer.
We set the hidden dimension as $d_h$.
For the $j$-th character, the local attention takes all the concatenated embeddings $ x_{j-\frac{k-1}{2}}, ...x_{j}, ... , x_{j+\frac{k-1}{2}} $ in the window as the input and outputs $k$ hidden vectors $ h_{j-\frac{k-1}{2}}, ..., h_{j}, ... , h_{j+\frac{k-1}{2}}$. 
The hidden vectors are calculated as follows:
\begin{equation}
	h_{m} = \alpha_{m} x_{m}
	\label{eq.localatt},
\end{equation}
where $ m \in \{j-\frac{k-1}{2},...,j+\frac{k-1}{2}\} $ and $\alpha_{m}$ is the attention weight, which is calculated as:
\begin{equation}
	\alpha_{m} = \frac{\exp{s(x_{j}, x_{m})}}{\sum_{n \in \{j-\frac{k-1}{2},...,j+\frac{k-1}{2}\}}\exp{s(x_{j}, x_{n})}}
	\label{eq.localattweight}.
\end{equation}
%For the $j$-th character, the local attention layer takes the concatenated embeddings $ x_{j-k/2}, ..., x_{j+k/2} $ as input and get $k$ hidden vectors $ h_{j-k/2}^{l}, ..., h_{j+k/2}^{l} $ for the CNN layer inputs as:
%\begin{equation}
%	h_{m}^{l} = \alpha_{m}^{l} x_{m}
%	\label{eq.localatt},
%\end{equation}
%where $ m \in \{j-k/2,...,j+k/2\} $ and the attention weight is calculated as:
%\begin{equation}
%	\alpha_{m}^{l} = \frac{\exp{s(x_{j}, x_{m})}}{\sum_{n \in \{j-k/2,...,j+k/2\}}\exp{s(x_{j}, x_{n})}}
%	\label{eq.localattweight}.
%\end{equation}
The score function $s$ is defined as follows:
\begin{equation}
	s(x_j, x_k) = v^{\top}\tanh(W_1 x_j + W_2 x_k)
	\label{eq.attscore},
\end{equation}
where $ v \in \mathbb{R}^{d_h} $ and $ W_1, W_2 \in \mathbb{R}^{d_h, d_e} $.
%where $ v^{l} \in \mathbb{R}^{d_h} $ and $ W_1^{l}, W_2^{l} \in \mathbb{R}^{d_h, d_e} $.

% The convolutional attention layer is used on the character sequence of each character to obtain its character representation $\mathbf{h}^c_i$, which is given by: 
% \begin{equation}
%   % \small
%     \mathbf{h}^c_{i} = \underset{t(i,1)\leq j\leq t(i,len(i))}{sum}
%     (\mathbf{W}^\top_\text{CNN}
%     \begin{bmatrix}
%     h_{j-\frac{k-1}{2}}\\
%     ...\\
%     h_{j+\frac{k-1}{2}}
%     \end{bmatrix}
%     + \mathbf{b}_\text{CNN})
%     \label{eq.cnn},
% \end{equation}
% where $\mathbf{W}_\text{CNN}$ and $\mathbf{b}_\text{CNN}$ are parameters and $sum$ denotes sum pooling.

The CNN layer contains $d_h$ kernels on a context window of $k$ tokens as:
\begin{equation}
	h_{j}^{c} = \sum_k[W^c*h_{j-\frac{k-1}{2}:j+\frac{k-1}{2}} + b^c]
	\label{eq.cnn},
\end{equation}
where $ W^c \in \mathbb{R}^{k \times d_h \times d_e} $ and $ b^c \in \mathbb{R}^{k \times d_h} $.
The operation $*$ denotes element-wise product and $h_{j-\frac{k-1}{2}:j+\frac{k-1}{2}}$ means a concatenation of the hidden states $ h_{j-\frac{k-1}{2}},...,h_{j+\frac{k-1}{2}} $, both of which are calculated at the first dimension.
Finally a sum-pooling is also conducted on the first dimension.

\subsection{BiGRU-CRF with Global Attention}
After extracting the local context features by convolutional attention layer, we feed them into a BiGRU-CRF based model to predict final label for each character.
%We feed all the features encoded from the convolutional attention layer above into a BiGRU-CRF based NER model to predict final entity type for each character.

The BiGRU layer is to model the sequential sentence information and calculated as follows:
\begin{equation}
	h_j^r = \text{BiGRU}(h_{j-1}^r, h_j^c; W^r, U^r)
	\label{eq.lstm},
\end{equation}
where $h_{j}^c$ is the output of convolutional attention layer, $h_{j-1}^r$ is the previous hidden state for the BiGRU layer, and $ W^r, U^r \in \mathbb{R}^{d_h \times d_h} $ are the parameters.
%which takes the hidden states of CNN layer as input and $ W^r, U^r \in \mathbb{R}^{d_h \times d_h} $ are the parameters for the input $h_j^c$ and the previous hidden state $h_{j-1}^r$, respectively.

%There is a problem that BiGRU-based models perform poorly on long sentences because of memory compression.
%are more computationally expensive to train and (这个不能提了，因为我们没有解决这个问题)
Then we apply a global self-attention layer to better handle the sentence-level information as:
\begin{equation}
	h_j^{g} = \sum_{s=1}^n \alpha_{j,s}^{g}h_s^r
	\label{eq.globalatt}
\end{equation}
where $ j = 1,...,\tau $ denotes all the characters in a sentence instance and $\alpha_{j,s}^{g}$ is calculated as:
\begin{equation}
	\alpha_{j,s}^{g} = \frac{\exp{s(h_{j}^r, h_{s}^r)}}{\sum_{n \in \{1,...,\tau\}}\exp{s(h_{j}^r, h_{n}^r)}}
	\label{eq.globalattweight}.
\end{equation}
The score function $s$ is similar to Equation~\ref{eq.attscore} with different parameters $ v^{g} \in \mathbb{R}^{d_h} $ and $ W_1^{g}, W_2^{g} \in \mathbb{R}^{d_h, d_h} $ instead.

Finally, a standard CRF layer is used on the top of the concatenation of the output of BiGRU layer and global attention layer, which is denoted as $H_\tau = [h_\tau^r; h_\tau^g]$. Given the predicted tag sequence $\mathbf{Y} = \{y_{1}, y_{2}, y_{3}, ..., y_{\tau}\}$, the probability of the ground-truth label sequence is computed by:
\begin{equation}
    P(\mathbf{Y}|\mathbf{X}) = \frac{exp(\sum_i(\mathbf{W}_{\text{CRF}}^{y_i} H_i + b_{\text{CRF}}^{(y_{i-1},y_i)}))}{\sum_{y'}exp(\sum_i(\mathbf{W}_{\text{CRF}}^{y'_{i}} H_i + b_{\text{CRF}}^{(y'_{i-1},y'_{i})}))},
\end{equation}
where $y'$ denotes an arbitary label sequence, $\mathbf{W}_{\text{CRF}}^{y_i}$ and $b_{\text{CRF}}^{(y_{i-1},y_i)}$ are trainable parameters. In decoding, we use Viterbi algorithm to get the predicted tag sequence. 

% Finally, we concatenate the output of the global self-attention layer with original BiGRU hidden to get the final layer output $ h_j = [h_j^r; h_j^g] $ and propose a linear chain CRF layer to predict the entity type for each character.
% The objective to optimize is the CRF likelihood defined as the following:
% \begin{equation}
% 	\small
% 	p(\mathbf{Y}|\mathbf{X};\theta) = \frac{1}{Z(\mathbf{X})} \prod_{j=1}^{|\mathbf{X}|}\exp \{ \theta \cdot f(y_j, y_{j-1}, \Phi(x_j)) \}
% 	\label{eq.crf},
% \end{equation}
% where $ \Phi(x_j) = h_j $ and $Z(x)$ is a normalization factor defined as:
% \begin{equation}
% 	\small
% 	Z(\mathbf{X}) = \sum_y \prod_{j=1}^{|\mathbf{X}|} \exp \{ \theta \cdot f(y_j, y_{j-1}, \Phi(x_j)) \}
% 	\label{eq.crfnorm}.
% \end{equation}
% The parameter $ \theta = \{ \theta_1, \theta_2 \} $ and objective function $f$ contain two parts as:
% \begin{equation}
% 	\small
% 	f = \sum_{j=1}^{|\mathbf{X}|} \{ p(y_j|y_{j-1};\theta_1) + p(y_j|\Phi(x_j);\theta_2) \}
% 	\label{eq.crf2parts}.
% \end{equation}

Till now we get the whole architecture which contains character-based, word-based, and sentence-based information altogether using multi-feature embeddings, CNN featurizers with local attention, and global self-attention mechanism.

\begin{table}[hb]
\centering
\small
\begin{tabular}{|c|c|c|c|c|}
    \hline
    \textbf{Dataset} & \textbf{Type} & \textbf{Train} & \textbf{Test} & \textbf{Dev} \\
    \hline
    \multirow{3}{*}{OntoNotes} & Sentences & 15.7k & 4.3k & 4.3k \\
    & Char & 491.9k & 208.1k & 200.5k \\
    & Entities & 13.4k & 7.7k & 6.95k \\
    \hline  
    \multirow{3}{*}{MSRA} & Sentences & 46.4k & 4.4k & - \\
    & Char & 2169.9k & 172.6k & - \\
    & Entities & 74.8k & 6.2k & - \\
    \hline
    \multirow{3}{*}{Weibo} & Sentences & 1.4k & 0.27k & 0.27k \\
    & Char & 73.8k & 14.8k & 14.5k \\
    & Entities & 1.89k & 0.42k & 0.39k \\
    \hline
    \multirow{3}{*}{Resume} & Sentences & 3.8k & 0.48k & 0.46k \\
    & Char & 124.1k & 15.1k & 13.9k \\
    & Entities & 1.34k & 0.15k & 0.16k \\
    \hline
\end{tabular}
\caption{Statistics of datasets}
\label{tb.statistic}
\end{table}

\subsection{Training}
% Assume that we have $K$ sentence-label pairs in total, $ \mathcal{D} = \{(\mathbf{X}_i, \mathbf{Y}_i)\}_{i=1}^K $.
% The final loss function is to maximize the following objective function:
% \begin{equation}
% 	\mathcal{O} = \sum_{i=1}^{K} p(\mathbf{Y}_i|\mathbf{X}_i;\theta)
% 	\label{eq.objective}.
% \end{equation}
For training, we exploit log-likelihood objective as the loss function. Given a set of training examples $\{(\mathbf{X}_i, \mathbf{Y}_i)\}|_{i=1}^K $, the loss function $L$ can be defined as follows:
\begin{equation}
    L = \sum_{i=1}^K log P(\mathbf{Y}_i|\mathbf{X}_i)
\end{equation}
In the training phase, at each iteration, we first shuffle all the training instances, and then feed them to the model with batch updates.
We use AdaDelta~\cite{zeiler2012adadelta} algorithm to optimize the final objective with all the parameters as described in Section~\ref{sec.expsetup}.

\section{Experiments}

To demonstrate the effectiveness of our proposed model, we did some experiments on Chinese NER datasets of different domains.
We will describe the details of datasets, setting and results in our experiments.
Standard precision (P), recall (R) and F1-score (F1) are used as evaluation metrics.
 
% \input{tables/statistics.tex}
% %\input{tables/hyperparas.tex}
% \input{tables/weibo.tex}
% %\input{tables/ontonotes.tex}
% \input{tables/resume.tex}
% \input{tables/msra.tex}
 
\subsection{Experimental Settings}
\label{sec.expsetup}
\textbf{Data} We use four datasets in our experiments.
%For news domain, we experiment on MSRA NER dataset of SIGHAN Bakeoff 2006 \cite{levow2006third}. 
For news domain, we experiment on OntoNotes 4 \cite{weischedel2011ontonotes} and MSRA NER dataset of SIGHAN Bakeoff 2006 \cite{levow2006third}. 
For the social media domain, we adopt the same annotated Weibo corpus as \cite{peng2015named} which is extracted from Sina Weibo \footnote{http://www.weibo.com/}.  
For more variety in test domains, we also use Chinese Resume dataset \cite{zhang2018chinese} collected from Sina Finance
\footnote{http://finance.sina.com.cn/stock/index.html}.
%MSRA NER dataset of SIGHAN Bakeoff 2006 \cite{levow2006third}.

Weibo dataset is annotated with four entity types: PER (Person), ORG (Organization), LOC (Location) and GPE (Geo-Political Entity), including named and nominal mentions.
Weibo corpus is already divided into training, development and test sets.
Chinese Resume dataset is annotated with eight types of named entities: CONT (Country), EDU (Educational Institution), LOC, PER, ORG, PRO (Profession), RACE (Ethnicity Background) and TITLE (Job Title). 
OntoNotes 4 dataset is annotated with four named entity categories: PER, ORG, LOC and GPE.
We follow the same data split method of \cite{che2013named} on OntoNotes.
For MSRA dataset, it only contains three annotated named entities: ORG, PER and LOC.
The development subset is not available in MSRA dataset.
\begin{table}[h]
\centering
\begin{tabular}{|c|c|c|c|}
    \hline
    \textbf{Models} & \textbf{NE} & \textbf{NM} & \textbf{Overall} \\
    \hline
    \cite[Peng and Dredze 2015]{peng2015named} & 51.96 & 61.05 & 56.05 \\
    \cite[Peng and Dredze 2016]{peng2016improving} & 55.28 & 62.97 & 58.99 \\
    \cite[He and Sun 2017]{he2017f} & 50.60 & 59.32 & 54.82 \\
    \cite[He and Sun 2017]{he2017unified} & 54.50 & 62.17 & 58.23 \\
    \cite[Cao et al. 2018]{cao2018adversarial} & 54.34 & 57.35 & 58.70 \\
    \cite[Zhang and Yang 2018]{zhang2018chinese} & 53.04 & 62.25 & 58.79 \\
    \hline  
    %TextFeaturizerChs & 42.11 & 49.40 & 50.07 \\
    %char feature embedding + gru + crf & 48.59 & 57.39 & 52.93\\
    %char feature embedding + gru + crf & 50.27 & 56.73 & 53.41\\
    %char feature embedding + gru + crf(256) & 48.47 & 60.12 & 54.10\\
    %char feature embedding + gru + crf(256) & 49.58 & 57.47 & 53.49\\
    Baseline & 49.02 & 58.80 & 53.80\\
    %char feature embedding + cnn + gru + crf & 51.61 & 60.99 & 56.25\\
    %char feature embedding + cnn + gru + crf & 53.26 & 59.15 & 56.15\\
    %char feature embedding + cnn + gru + crf(256) & 54.10 & 57.69 & 55.89\\
    %char feature embedding + cnn + gru + crf(256) & 53.61 & 58.40 & 55.93\\
    Baseline + CNN & 53.86 & 58.05 & 55.91\\
    %char feature embedding + att-cnn + gru + global attention + crf & - & - & 60.39\\
   % char feature embedding + att-cnn + gru + global attention + crf & 54.64 & 63.19& 58.77\\
   % char feature embedding + att-cnn + gru + global attention + crf & 53.83 & 62.22 & 58.19\\
    CAN Model & \textbf{55.38} & \textbf{62.98} & \textbf{59.31}\\
    
    \hline
\end{tabular}
\caption{Weibo NER results}
\label{tb.weibo}
\end{table} 

The detail statistic information of our datasets is shown in Table \ref{tb.statistic}.

%\textbf{Segmentation Feature}
%Gold segmentation is already available for the OntoNotes 4 dataset in the training section, test section and development section. 
%Gold segmentation is already available for the MSRA dataset in the training section.
%For MSRA dataset, gold segmentation is available for the training section. 
Gold segmentation is unavailable for Weibo dataset, Chinese Resume datasets and MSRA test sections. 
We followed \cite{zhang2018chinese} to automatically segment Weibo dataset, Chinese Resume dataset and MSRA test sections using the model of \cite{yang2017neural}.
%
%\textbf{Labeling Scheme} 
We treat NER as a sequential labeling problem and adopt BIOES tagging style in this paper since it has been shown that models using BIOES are remarkably better than BIO \cite{yang2018Design}.
%In BIOES tagging scheme, O means that the character is not a part of Named Entity (NE); B-X represents the beginning character of a NE type X; I denotes that the character is inside of a NE type X; E means the end character of a NE; and S stands for a NE that only contains one character. 
%In BIOES tagging scheme, O means that the character is not a part of Named Entity (NE); B/I/E/S-X represents the beginning/inside/ending/single character of a NE type X;

%\input{tables/statistics.tex}

\textbf{Hyper-parameter settings}
For hyper-parameter configuration, we adjust them according to the performance on the development set of Chinese NER task. 
We set the character embedding size, hidden sizes of CNN and Bi-GRU  to 300 dims. 
After comparing to experimental results with different window sizes of CNN, we set the window size as 5. 
%that it hardly affects the final quality.
%Our segmentation feature size is set to 20. 
%Dropout rate is 0.45. 
Adadelta is used for optimization, with an initial learning rate of 0.005. 
%The batch size of Weibo dataset and Chinese Resume dataset are set to 5, while the batch size of OntoNotes 4 is 100. 
The character embeddings used in our experiments are from \cite{li2018Analogical}, which is trained by Skip-Gram with Negative Sampling (SGNS) on Baidu Encyclopedia.

\subsection{Experimental Results}
In this section, we will give the experimental results of our proposed model and previous state-of-the-art methods on Weibo dataset, Chinese Resume dataset, OntoNotes 4 dataset and MSRA dataset, respectively.
We propose two baselines and a CAN model. 
In the experiment results table, we use Baseline to represent the BiGRU + CRF model and Baseline + CNN to indicate CNN + BiGRU + CRF.

\subsubsection{Weibo Dataset}
We compare our proposed model with the latest models on Weibo dataset. 
Table \ref{tb.weibo} shows the F1-scores for named entities (NE), nominal entities (NM, excluding named entities) and both (Overall). 
We observe that our proposed model achieves state-of-the-art performance.

Existing state-of-the-art systems include \cite{peng2016improving}, \cite{he2017unified}, \cite{cao2018adversarial} and \cite{zhang2018chinese}, who leverage rich external data like cross-domain data, semi-supervised data and lexicon or joint train NER task with Chinese Word Segmentation (CWS) task.\footnote{The results of \cite{peng2015named,peng2016improving} are taken from \cite{peng2017supplementary}}
In the first block of Table \ref{tb.weibo}, we report the performance of the latest models. 
The model that jointly train embeddings with NER task proposed by \cite{peng2015named} achieves F1-score of 56.05\% on overall performance. 
The model \cite{peng2016improving} that jointly train CWS task improves the F1-score to 58.99\%. 
\cite{he2017unified} propose a unified model to exploit cross-domain and semi-supervised data, which improves the F1-score from 54.82\% to 58.23\% compared with the model proposed by \cite{he2017f}. 
\cite{cao2018adversarial} use an adversarial transfer learning framework to incorporate task-shared word boundary information from CWS task and achieves F1-score of 58.70\%. 
\cite{zhang2018chinese} leverage a lattice structure to integrate lexicon information into their model and achieve F1-score of 58.79\%.

In the second block of Table \ref{tb.weibo}, we give the results of our baselines and proposed models. Our baseline Bi-GRU + CRF achieves a F1-score of 53.80\% and adding a normal CNN layer as featurizer improve F1-score to 55.91\%. Replacing normal CNN with our convolutional attention layer will significantly improve the F1-score to 59.31\%, which is the highest result among existing models. The improvement demonstrates the effectiveness of our proposed model.

\begin{table}[thbp]
\centering
\begin{tabular}{|c|c|c|c|}
    \hline
    \textbf{Models} & \textbf{P} & \textbf{R} & \textbf{F1} \\
    \hline
    \cite[Zhang and Yang 2018]{zhang2018chinese}$^1$ & 94.53 & 94.29 & 94.41 \\
    \cite[Zhang and Yang 2018]{zhang2018chinese}$^2$ & 94.07 & 94.42 & 94.24 \\
    \cite[Zhang and Yang 2018]{zhang2018chinese}$^3$ & 94.81 & 94.11 & 94.46 \\
    \hline  
    %char feature embedding + gru + crf & 93.62 & 93.56 & 93.59\\
    %char feature embedding + gru + crf & 94.16 & 94.05 & 94.11\\
    
    %char feature embedding + gru + crf & 93.80 & 93.80 & 93.80\\
    %char feature embedding + gru + crf & 93.62 & 93.68 & 93.65\\
    Baseline & 93.71 & 93.74 & 93.73\\
    %char feature embedding + cnn + gru + crf & 94.56 & 94.91 & 94.73\\
    %char feature embedding + cnn + gru + crf & 94.74 & 95.03 & 94.89\\
    %char feature embedding + cnn + gru + crf & 94.09 & 94.79 & 94.44\\
    %char feature embedding + cnn + gru + crf & 94.62 & 94.91 & 94.76\\
    Baseline + CNN & 94.36 & \textbf{94.85} & 94.60\\ 
    %char feature embedding + att-cnn + gru + global attention + crf & 95.26 & 94.85 & 95.05\\
    %char feature embedding + att-cnn + gru + global attention + crf & 94.85 & 94.91 & 94.89\\
    %char feature embedding + att-cnn + gru + global attention + crf & 95.25 & 94.72 & 94.99\\
    CAN Model & \textbf{95.05} & 94.82 & \textbf{94.94}\\
    
    \hline
\end{tabular}
\caption{Results on Chinese Resume Dataset. For models proposed by \cite{zhang2018chinese}, $1$ represents the char-based LSTM model, $2$ indicates the word-based LSTM model and $3$ is the Lattice model. }
\label{tb.resume}
\end{table} 
\subsubsection{Chinese Resume Dataset}
The Chinese Resume test results are shown in Table \ref{tb.resume}. 
\cite{zhang2018chinese} release Chinese Resume dataset and achieve F1-score of 94.46\% with lattice structure incorporating additional lexicon information. 
It can be seen that our proposed baseline (CNN + Bi-GRU + CRF) outperforms \cite{zhang2018chinese} with F1-score of 94.60\%. Adding our convolutional attention leads a further improvement and achieves state-of-the-art F1-score of 94.94\%, which demonstrates the effectiveness of our proposed model.

\subsubsection{OntoNotes Dataset}
Table \ref{tb.onto4} shows the comparisons on OntoNotes 4 dataset.
\footnote{In Table \ref{tb.onto4} and \ref{tb.msra}, we use $\ast$ to denote a model with external labeled data for semi-supervised learning. $\dagger$ denotes that the model use external lexicon data. \cite{zhang2018chinese} with $\ddagger$ is the char-based model in the paper.}
In the first block, we list the performance of previous methods for Chinese NER task on OntoNotes 4 dataset. \cite{yang2017combining} propose a model combining neural and discrete feature, e.g., POS tagging features, CWS features and orthographic features, improving the F1-score from 68.57\% to 76.40\%. Leveraging bilingual data, \cite{che2013named} and \cite{wang2013effective} achieves F1-score of 74.32\% and 73.88\% respectively.
\cite{zhang2018chinese}$^\ddagger$ is the character-based model with bichar and softword.

In the second block of Table \ref{tb.onto4}, we give the results of our baselines and proposed models.
Consistent with observations on the Weibo and Resume datasets, our Convolutional Attention layer leads an increment of F1-score and our proposed model achieves a competitive F1-score of 73.64\% among character-based model without using external data.

%\begin{figure}[t]
%    \centering
%    \includegraphics[width=\linewidth]{figures/local_att.png}
%    \caption{Convolutional attention.}
%    \label{fg.convatt}
%\end{figure}

%\begin{figure}[t]
%    \centering
%    \includegraphics[width=\linewidth]{figures/global_att.png}
%    \caption{Global attention.}
%    \label{fg.gloatt}
%\end{figure}
\begin{table}[t]
\centering
\begin{tabular}{|c|c|c|c|}
    \hline
    \textbf{Models} & \textbf{P} & \textbf{R} & \textbf{F1} \\
    \hline
    \cite[Yang et al. 2016]{yang2017combining} & 65.59 & 71.84 & 68.57 \\
    \cite[Yang et al. 2016]{yang2017combining}$^\ast$ & 72.98 & \textbf{80.15} & \textbf{76.40} \\
    \cite[Che et al. 2013]{che2013named}$^\ast$ & \textbf{77.71} & 72.51 & 75.02 \\
    \cite[Wang et al. 2013]{wang2013effective}$^\ast$ & 76.43 & 72.32 & 74.32 \\
    \cite[Zhang and Yang 2018]{zhang2018chinese}$^\dagger$ & 76.35 & 71.56 & 73.88 \\
    \cite[Zhang and Yang 2018]{zhang2018chinese}$^\ddagger$ & 74.36 & 69.43 & 71.81 \\
    \hline
    %TextFeaturizerChs & 70.45 & 56.09 & 62.45 \\
    %char feature embedding + gru + crf(batch=400) & 71.24 & 63.37 & 67.07\\
    Baseline & 70.67 & 71.64& 71.15\\
    %char feature embedding + cnn + gru + crf(batch=400) &71.73 & 67.62 & 69.61\\
    Baseline + CNN &72.69 & 71.51 & 72.10\\
    %char feature embedding + att-cnn + gru + global attention + crf(batch=400) & 70.26 & 68.69 & 69.47 \\
    %char feature embedding + att-cnn + gru + global attention + crf(batch=100) & 71.82 & 71.71 & 71.76 \\
    %char feature embedding + att-cnn + gru + global attention + crf(batch=100) & 72.63 & 71.37 & 71.99 \\
    %char feature embedding + att-cnn + gru + global attention + crf(batch=100) & 71.06 & 72.61 & 71.82 \\
    
    % CAN Model & 73.63 & 70.82 & 72.20 \\
    CAN Model & 75.05 & 72.29 & 73.64 \\
    
    %char feature embedding + att-cnn + gru + global attention + crf(train+dev,batch=400,150) & 79.42 & 76.63 & 78.00 \\
    %char feature embedding + att-cnn + gru + global attention + crf(train+dev,batch=400,看dev) & 78.80 & 77.77 & 78.28\\
    %char feature embedding + att-cnn + gru + global attention + crf(batch=50) + lexicon & 72.63 & 71.37 & 73.02 \\
    \hline
\end{tabular}
\caption{Results on OntoNotes}
\label{tb.onto4}
\end{table} 

\subsubsection{MSRA Dataset}
Table \ref{tb.msra} shows comparisons on MSRA dataset. 
In the first block, we give the performance of previous methods for Chinese NER task on MSRA dataset.
\cite{chen2006chinese}, \cite{zhang2006word} and \cite{zhou2013chinese} leverage rich hand-crafted features and \cite{lu2016multi} exploit multi-prototype embeddings features for Chinese NER task.
\cite{dong2016character} introduce radical features into LSTM-CRF.
\cite{cao2018adversarial} make use of Adversarial Transfer Learning and global self-attention to improve performance.
\cite{zhoufive} propose a character-based CNN-BiLSTM-CRF model to incorporate stroke embeddings and generate n-gram features.
\cite{zhang2018chinese} introduce a lattice structure to incorporate lexicon information into the neural network, which actually includes word embedding information. 
Although the model achieves state-of-the-art F1-score of 93.18\%, it leverages external lexicon data and the result is dependent on the quality of the lexicon. 

In the second block, we list baselines and proposed model. It can seen that our Baseline + CNN is already outperform most previous methods. Compared with the state-of-the-art model proposed by \cite{zhang2018chinese}, our char-based method achieves competitive F1-score of 92.97\% without additional lexicon data and word embedding information. Our CAN model achieves state-of-the-art result among the character-based models.

\begin{table}[tbp]
\centering
\begin{tabular}{|c|c|c|c|}
\hline
    \textbf{Models} & \textbf{P} & \textbf{R} & \textbf{F1} \\
    \hline
    \cite[Chen et al. 2006]{chen2006chinese} & 91.22 & 81.71 & 86.20 \\
    \cite[Zhang et al.2006]{zhang2006word}$^\ast$ & 92.20 & 90.18 & 91.18 \\
    \cite[Zhou et al.2013]{zhou2013chinese} & 91.86 & 88.75 & 90.28 \\
    \cite[Lu et al. 2016]{lu2016multi} & - & - & 87.94 \\
    \cite[Dong et al. 2016]{dong2016character} & 91.28 & 90.62 & 90.95 \\
    \cite[Cao et al. 2018]{cao2018adversarial} & 91.30 & 89.58 & 90.64 \\
    \cite[Zhou et al. 2018 ]{zhoufive} & 92.04 & 91.31 & 91.67 \\
    \cite[Zhang and Yang 2018]{zhang2018chinese} & \textbf{93.57} & \textbf{92.79} & \textbf{93.18} \\
    \hline
    %TextFeaturizerChs & 88.50 & 80.01 & 84.04 \\
    Baseline & 92.54 & 88.20 & 90.32 \\
    Baseline + CNN & 92.57 & 92.11 & 92.34 \\
    CAN Model & \textbf{93.53} & 92.42 & \textbf{92.97} \\
    \hline
\end{tabular}
\caption{Results on MSRA dataset}
\label{tb.msra}
\end{table}

 \begin{figure*}[th]
     \centering
     \begin{subfigure}[c]{0.5\textwidth}
         \centering
         \includegraphics[width=\linewidth]{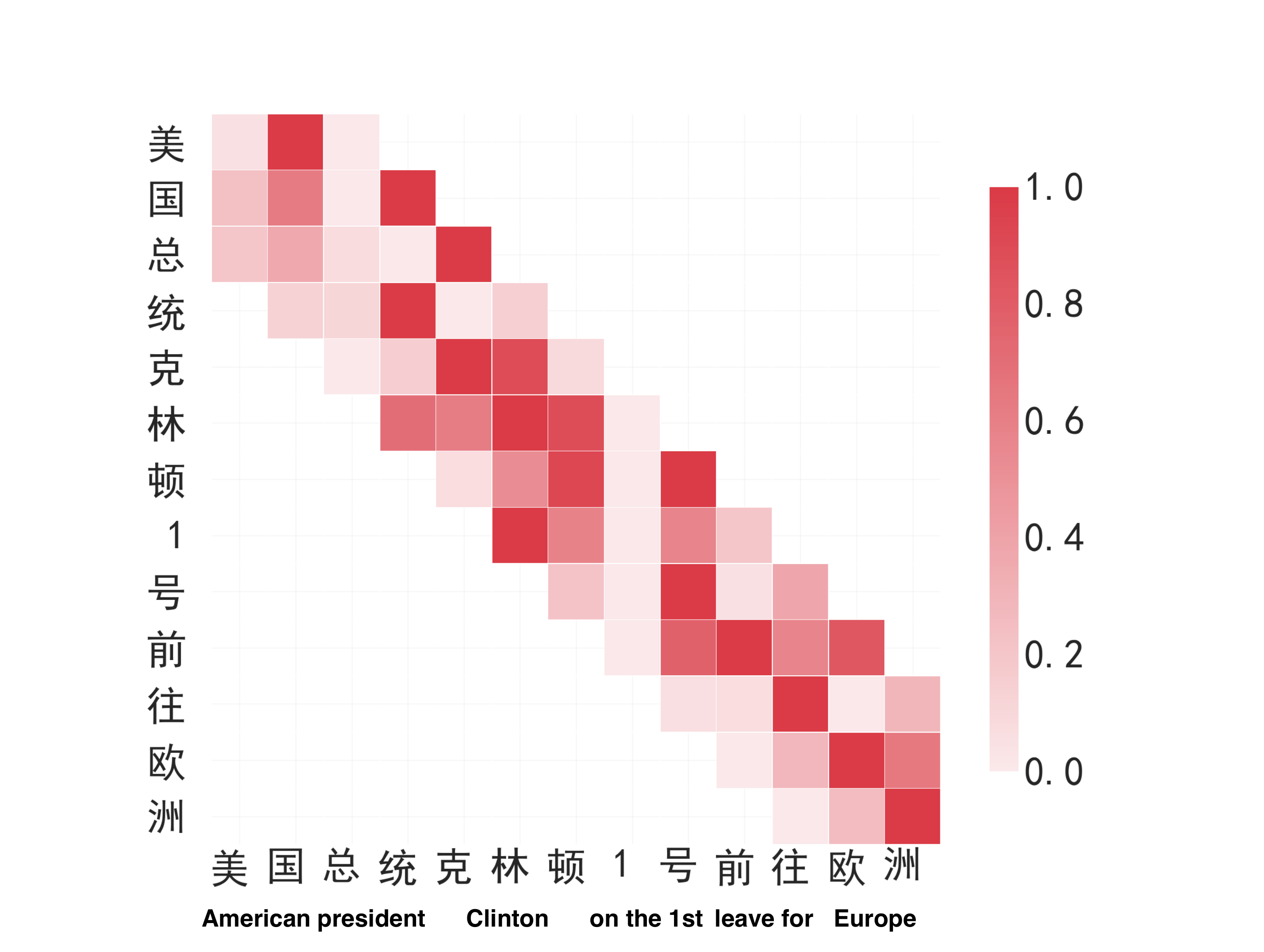}
         \caption{Local attention.}
         \label{fg.localatt}
     \end{subfigure}%
     ~ 
     \begin{subfigure}[c]{0.5\textwidth}
         \centering
         \includegraphics[width=\linewidth]{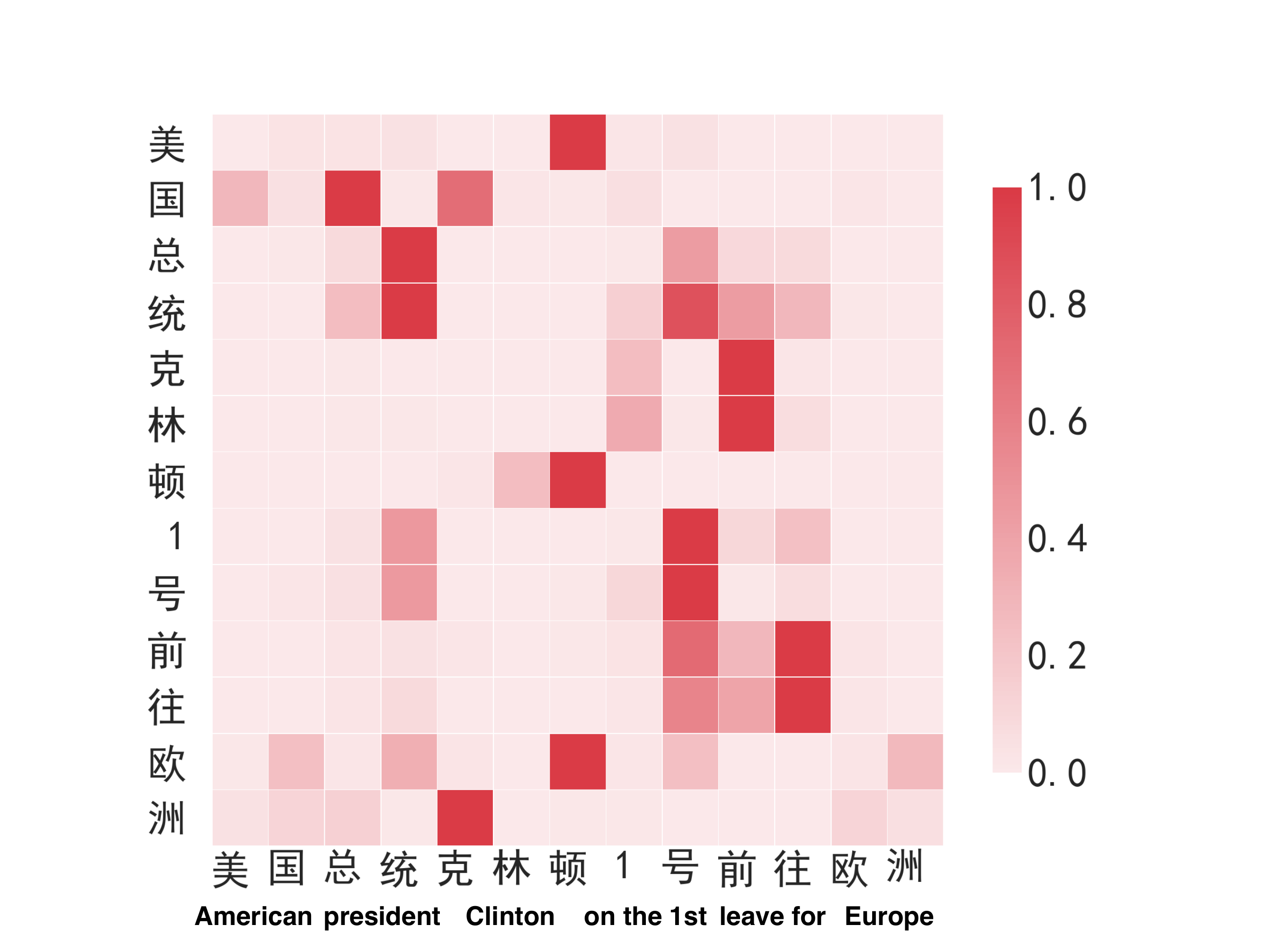}
         \caption{Global self-attention.}
         \label{fg.globalatt}
     \end{subfigure}
     \caption{Attention visualization. The left part shows the normalized Convolutional Attention weights in each window in a sentence. The right part indicates the global self-attention weights among the whole sentence. For both pictures, the x-axis represents the context while the y-axis on behalf of the query in the attention mechanism. }
 \end{figure*}

\subsection{Discussion}
\subsubsection{Effectiveness of Convolutional Attention and Global Self-Attention}
Shown by Table \ref{tb.weibo}, \ref{tb.resume} and \ref{tb.msra}, the performance of our proposed model demonstrates the effectiveness of the Convolutional Attention Network. To better evaluate the effect of the Attention Mechanism, we visualize the normalized attention weights $\alpha_{m}^{l}$ for each window from Eq. \ref{eq.localattweight}, as in Figure \ref{fg.localatt}. 
Each row of the matrix represents location attention weights in each window. 
For example, the third row indicates that the relationship between center character “总” and contexts “美 国 总 统 克”. 
We can see from the Figure \ref{fg.localatt} that the word-level features can be extracted through the local attention. 
In the context, the center character “美” tends to have a stronger connection with its related character “国”, which means they have a higher possibility of consisting of a Chinese word “美国 (American)”. 
%We can find the same pattern in Figure \ref{fg.localatt}.
%For example,
Also for characters “克”,“林” and “顿” tend to have a strong connection because “克林顿” means “Clinton”. 
Character “欧” and “洲” have strong connection seen from the Figure \ref{fg.localatt} because “欧洲” represents “Europe” in Chinese.
Therefore, both experiments results and visualization verifies that the Convolutional Attention is effective for obtaining the phrase information between adjacent characters. 

In Figure \ref{fg.globalatt}, we visualize the global self-attention matrix. 
From the picture, we can find that global self-attention can capture the sentence context information from the long-distance relationship of words to overcome the limitation of Recurrent Neural Network.
For the word “克林顿 (Clinton)”, the global self-attention learns the dependencies with “前往 (leave for)” and “1号 (on the 1st)”. Distinguished by the color,  “克林顿 (Clinton)” has a stronger connection with “前往 (leave for)” than with “1号 (on the 1st)”, which accords with the expectation that the predicate in the sentence provides more information to the subject than the adverbial of time. 

%It demonstrates that the self-attention is effective for Chinese NER task.

%\subsubsection{Case Study}

\subsubsection{Results Analysis}
Our proposed model outperforms previous work on Weibo and Chinese Resume dataset and gains competitive results on MSRA and OntoNotes 4 datasets without using any external resources. 
The experiments results demonstrate the effectiveness of our proposed model, especially among the char-based model. 
The performance improvement after adding Convolutional Attention Layer and Global Attention Layer verifies that our model can capture the relationship between character and its local context, as well as the relationship between word and global context. 
However, although we can obtain comparable results compared with other models without external resources, we find that our model performs relatively unsatisfying in the OntoNotes 4 dataset. It may be explained by the reason that discrete features and external resources like other labeled data or lexicons have a more positive influence on this dataset when the model cannot learn enough information from the training set.
%because the formal expressions in the corpus of the news domain can be easily detected through external lexicons.   

%\input{tables/ontonotes.tex}

%\input{tables/resume.tex}

%\noindent\textbf{msra}
%\input{tables/msra.tex}
%\input{sections/analysis.tex}
\section{Related Work}

\subsection{Neural Network Models}
It has been shown that neural networks, such as LSTM and CNN, can outperform conventional machine learning methods without requiring handcrafted features.
\cite{collobert2011natural} applied a CNN-CRF model and gained competitive results to the best statistical models. 
More recently, the LSTM-CRF architecture has been used on NER tasks.
\cite{huang2015bidirectional} employed BiLSTM to extract word-level context information and \cite{lample2016neural} futher introduced hierarchy structure by incorporating BiLSTM-based character embeddings.
Many works integrating word-level information and character-level information have been found to achieve good performance \cite{dos2015boosting,chiu2016named,ma2016end,lample2016neural,chen2019grn}.
External knowledge has also been exploited for NER tasks.
To utilize character-level knowledge, character-level pre-trained \cite{peters2017semi} and co-trained \cite{AAAI1817123} neural language models were introduced. 
Recently, many works exploit learning pre-trained language representations with deep language models to improve the performance of downstream NLP tasks, such as ELMo \cite{peters2018deep} and BERT \cite{devlin2018bert}.

\subsection{Attention Mechanism related Models}
Recently, Attention Mechanism has shown a very good performance on a variety of tasks including machine translation, machine comprehension and related NLP tasks \cite{vaswani2017attention,seo2016bidirectional,tan2017deep}.
In language understanding task, \cite{shen2017disan} exploit self-attention to learn long range dependencies.
\cite{rei2016attending} proposed the model employing an attention mechanism to combine the character-based representation with the word embedding instead of simply concatenating them.
This method allows the model to dynamically decide which source of information to use for each word, and therefore outperforming the concatenation method used in previous work.
\cite{zhang2018adaptive} use pictures in Tweets as external information through an adaptive co-attention network to decide whether and how to integrate the images into the model.
The method can only apply to websites like Tweets which has text-related images, but the resources like that are insufficient.
\cite{tan2018deep} and \cite{cao2018adversarial} employ self-attention to directly capture the global dependencies of the inputs for NER task and demonstrate the effectiveness of self-attention in Chinese NER task.

%To our best knowledge, we are the first to apply attention mechanism on languages without explicit word separators (like Chinese) and achieve state-of-the-art on Chinese datasets.(我们现在好像不是first了)

\subsection{Chinese NER}
Multiple previous works tried to address the problems that the Chinese language doesn't have explicit word boundaries.
Traditional models depended on hand-crafted features and CRFs-based models \cite{he2008chinese,mao2008chinese}. 
Character-based LSTM-CRF model was applied to Chinese NER to utilize both character-level and radical-level representations \cite{dong2016character}.
\cite{peng2015named} applied character positional embeddings and proposed a jointly trained model for embeddings and NER.
To better integrate word boundary information into Chinese NER model, \cite{peng2016improving} co-trained NER and word segmentation to improve both tasks.
\cite{he2017unified} unified cross-domain learning and semi-supervised learning to obtain information from out-of-domain corpora and in-domain unannotated texts.
Instead of performing word segmentation first, \cite{zhang2018chinese} constructed a word-character lattice by matching words in texts with a lexicon to avoid segmentation errors. 
\cite{cao2018adversarial} used the adversarial network to jointly train Chinese NER task and Chinese Word Segmentation task to extract task-shared word boundary information.
\cite{yang2018adversarial} leveraged character-level BiLSTM to extract higher-level features from crowd-annotations.
\section{Conclusion}
In this paper, we propose a Convolutional Attention Network model to improve Chinese NER model performance and preclude word embedding and additional lexicon dependencies, that makes the model more efficient and robust.
In our model, we implement local-attention CNN and Bi-GRU with the global self-attention structure to capture word-level features and context information with char-level features.
Experiments show that our model outperforms the state-of-art systems on the different domain datasets.

For future works, we would like to study how to joint learning word segmentation and NER tasks to further reduce constraints.

\section*{Acknowledgement}
We thank our colleagues Haoyan Liu, Zijia Lin, as well as the anonymous reviewers for their valuable feedback.

\end{CJK}

\bibliography{main}
\bibliographystyle{IEEEtran}

\end{document}